\documentclass[10pt,twocolumn,letterpaper]{article}

\usepackage{iccv}
\usepackage{times}
\usepackage{epsfig}
\usepackage{graphicx}
\usepackage{amsmath}
\usepackage{amssymb}

\usepackage{booktabs} %
\usepackage{amsbsy}
\usepackage{xcolor}
\usepackage{color}
\usepackage{xspace}
\usepackage{algorithm}
\usepackage{algorithmic}
\usepackage{multirow}
\usepackage{multicol}
\usepackage{float}
\usepackage{relsize}
\usepackage{caption}

\usepackage{bbm}
\usepackage{soul}
\usepackage{lipsum}

\definecolor{darkgreen}{rgb}{0,0.45,0}
\definecolor{darkred}{rgb}{0.5,0,0}
\definecolor{purple}{rgb}{0.5,0,0.5}

\newcommand{\ours}{$\text{MAL}$\xspace}
\newcommand{\bdd}{$\text{BDD100K}$\xspace}
\newcommand{\imagenet}{$\text{ImageNet}$\xspace}
\newcommand{\cifar}{$\text{CIFAR100}$\xspace}
\newcommand{\imbalancedcifar}{$\text{CIFAR100-Imbalanced}$\xspace}
\newcommand{\inat}{$\text{iNaturalist2018}$\xspace}
\newcommand{\city}{$\text{Cityscapes}$\xspace}

\usepackage[pagebackref=true,breaklinks=true,letterpaper=true,colorlinks,bookmarks=false]{hyperref}

 \iccvfinalcopy %

\def\iccvPaperID{9653} %

\ificcvfinal\fi

\begin{document}

\title{Minimax Active Learning}

\author{
	Sayna Ebrahimi\\
	UC Berkeley
\and
William Gan\\
UC Berkeley
\and 
Dian Chen\\
UC Berkeley
\and 
Giscard Biamby\\
UC Berkeley
\and 
Kamyar Salahi\\
UC Berkeley
\and
Michael Laielli\\
UC Berkeley
\and 
Shizhan Zhu\\
UC Berkeley
\and 
Trevor Darrell\\
UC Berkeley\\
\and
{\tt\small \{sayna, wjgan, dian, gbiamby, kam.salahi, laielli, shizhan\_zhu,  trevordarrell\}@berkeley.edu}
}

\maketitle
\ificcvfinal\thispagestyle{empty}\fi

\begin{abstract}
	Active learning aims to develop label-efficient algorithms by querying the most representative samples to be labeled by a human annotator. Current active learning techniques either rely on model uncertainty to select the most uncertain samples or use clustering or reconstruction to choose the most diverse set of unlabeled examples. While uncertainty-based strategies are susceptible to outliers, solely relying on sample diversity does not capture the information available on the main task. In this work, we develop a semi-supervised minimax entropy-based active learning algorithm that leverages both uncertainty and diversity in an adversarial manner. Our model consists of an entropy minimizing feature encoding network followed by an entropy maximizing classification layer. This minimax formulation reduces the distribution gap between the labeled/unlabeled data, while a discriminator is simultaneously trained to distinguish the labeled/unlabeled data. The highest entropy samples from the classifier that the discriminator predicts as unlabeled are selected for labeling. We evaluate our method on various image classification and semantic segmentation benchmark datasets and show superior performance over the state-of-the-art methods. \footnote{Project page: \url{https://people.eecs.berkeley.edu/~sayna/mal.html}}
\end{abstract}

\section{Introduction}
The outstanding performance of modern computer vision systems on a variety of challenging problems~\cite{feng2019computer} is empowered by several factors: recent advances in deep neural networks~\cite{dosovitskiy2020image}, increased computing power~\cite{djolonga2020robustness}, and an abundant stream of data~\cite{kolesnikov2019big}, which is often expensive to obtain. Active learning focuses on reducing the amount of human-annotated data needed to obtain a given performance by iteratively selecting the most \textit{informative} samples for annotation.

\begin{figure*}[t]
	\begin{center}
		\includegraphics[width=0.9\linewidth]{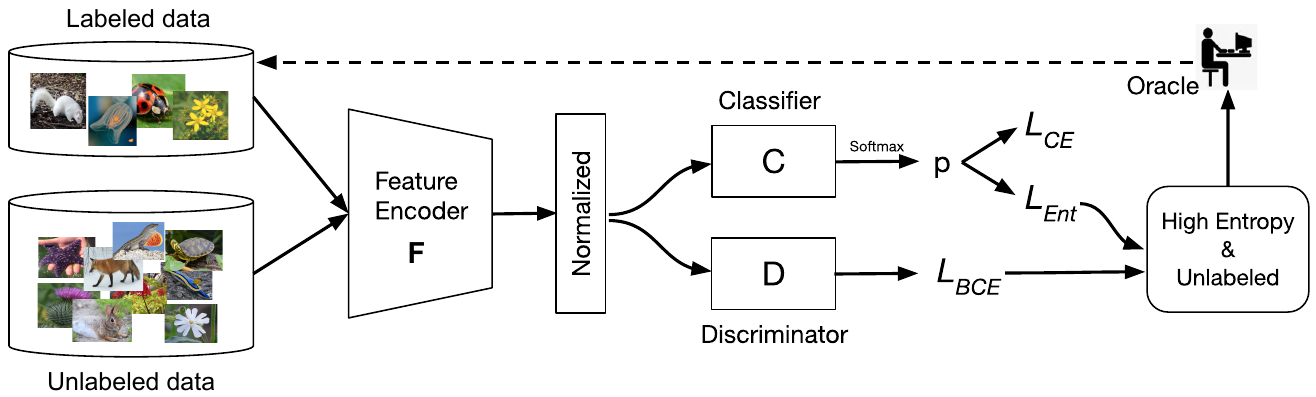}
	\end{center}
	\caption{The \ours pipeline: both labeled and unlabeled data are input into a common feature encoder $F$, followed by a normalization layer and a cosine similarity-based classifier $C$ which is trained to maximize entropy on unlabeled data, whereas $F$ is trained to minimize it. We also train a discriminator $D$ (a binary classifier) to predict \textit{labeledness} of the data. Samples with highest entropy from the classifier that the discriminator predicts as unlabeled are selected for labeling.}
	\label{fig:teaser}
\end{figure*}

Notably, a large research effort in active learning, implicitly or explicitly, leverages the uncertainty associated with the outcomes on the main task for which we are trying to annotate the data~\cite{wang2014label,kirsch2019batchbald,dbal,gorriz2017costeffectivemelanoma,wang2017cost, beluch2018ensemble}. These approaches are vulnerable to outliers and are prone to annotate redundant samples in the beginning because the task model is equally uncertain on almost all the unlabeled data, regardless of how representative the data are for each class~\cite{vaal}. 
As an alternative direction, \cite{vaal} introduced a \textit{task-agnostic} approach in which they applied reconstruction as an unsupervised task on the labeled and unlabeled data, which allowed for training a discriminator adversarially to predict whether a given sample should be labeled.  

Task-agnostic approaches achieve strong performance on a variety of datasets~\cite{vaal} and can select new samples for labeling without requiring updated annotations from the labeler, i.e. because task-agnostic approaches do not use the annotations.
But by ignoring the annotations, task-agnostic approaches disregard how similar the unlabeled samples are to the labeled sample classes. As a result, the samples selected by the task-agnostic discriminator can all come from the same class, even if there is already a high proportion of labeled samples from that class. %

In this paper, we empirically show that the performance increase from using all data as well as all labels can be significant in real-world settings where datasets are imbalanced and poor performance on a specific class can be highly correlated with having fewer labeled samples for that class. Therefore, ranking the unlabeled data with respect to its entropy in addition to its similarity to the existing labeled data, which we refer to as \textit{labeledness} score, can guide the acquisition function to not only query samples that \textit{do not look similar to what has been seen before}, but also rank them based on \textit{how far they are from the predicted class prototypes}.

This paper introduces a semi-supervised pool-based task-aware active learning algorithm that combines the best of both directions in active learning literature. Our approach, called Minimax Active Learning (\ours), learns a representation on both labeled and unlabeled data and queries samples based on their \textit{diversity} as well as their \textit{uncertainty} scores obtained by a distance-based classifier trained on the main task. The diversity or \textit{labeledness} of the data is predicted by a discriminator that is trained to distinguish between the labeled and unlabeled features generated by an encoder that optimizes a minimax objective to reduce intra-class variations in the dataset.  Given two sets of labeled and unlabeled data, it is difficult to train a classifier that can distinguish between them without overfitting to the dominant set (unlabeled pool). Our key idea is to learn a latent space representation that is adversarially trained to become discriminative while we minimize the expected loss on the labeled data. 

Figure \ref{fig:teaser} illustrates our model architecture. In particular, \ours has three major building blocks: \textbf{(i)} a feature encoder ($F$) that attempts to minimize the entropy of the unlabeled data for better clustering. \textbf{(ii)} a distance-based classifier ($C$) that adversarially learns per-class weight vectors as prototypes and attempts to maximize the entropy of the unlabeled data \textbf{(iii)} a discriminator ($D$) that uses features generated by $F$ to predict which pool each sample belongs to. Inspired by recent advances in few-shot learning, we use a distance-based classifier to learn per-class weighted vectors as prototypes (similar to prototypical~\cite{prototypical} and matching networks~\cite{matchingnets}) using cosine similarity scores~\cite{baseline++}. To leverage the unlabeled data, we perform semi-supervised learning using entropy as our optimization objective which is easy to train and does not require labels. 

While the feature extractor minimizes the entropy by pushing the unlabeled samples to be clustered near the prototypes, the classifier increases the entropy of the unlabeled samples, and hence makes them similar to all class prototypes. On the other hand, they both learn to classify the labeled data using a cross-entropy loss, enhancing the feature space with more discriminative features. We train a binary discriminator on this space to predict \textit{labeledness} for each example and at the end samples which (i) are predicted as unlabeled with high confidence by the discriminator and (ii) achieved high entropy by the classifier $C$ will be sent to the oracle. 

While distance-based classification~\cite{metriclearning,lowshot,baseline++} and minimax entropy for image  classification~\cite{jaynes, kullback,minimax,mme,mmen} have been studied before, to the best of our knowledge, our work is the first to use them together for active learning. we will demonstrate our semi-supervised adversarial entropy optimization on labeled and unlabeled data can significantly outperform the fully supervised active learning strategies that use uncertainty, diversity, or random sampling. This work is also the first to show semi-supervised learning with minimax entropy for semantic segmentation. Our results demonstrate significant performance improvements on a variety of image classification benchmarks such as \cifar, \imagenet, and \inat and semantic segmentation using \city and \bdd driving datasets.

\section{Related Work}
\noindent \textbf{Active learning:}
Recent work in active learning can be divided into two sets of approaches: query-synthesizing (generative) methods or query-acquiring (pool-based) methods. Query-synthesizing methods produce informative synthetic samples using generative models \cite{mahapatra2018efficient, adversarialsamplingactivelearning, zhu2017generative, pmlr-v97-tran19a} while pool-based approaches focus on selecting the most informative samples from a data pool. We focus chiefly on the latter line of research.

Pool-based methods have been shown both theoretically and empirically to achieve better performance than Random sampling~\cite{settles2014active,freund1997selective,gilad2006query}. Among pool-based methods, there exist three major categories: uncertainty-based approaches~\cite{gorriz2017costeffectivemelanoma, wang2017cost, beluch2018ensemble}, representation-based (or diversity) approaches~\cite{sener2018coreset,wang2014label,vaal}, and hybrid approaches~\cite{suggestiveannotation,preclustering,ash2020deep}. Pool-based sampling strategies have been studied widely with early works such as information-theoretic methods~\cite{mackay1992information}, ensemble methods~\cite{mccallumzy1998employing,freund1997selective} and uncertainty heuristics~\cite{tong2001support,li2013adaptive} surveyed by Settles~\cite{activelearningsurvey}.

Uncertainty-based pool-based methods have been studied in both Bayesian~\cite{dbal, kirsch2019batchbald} and non-Bayesian settings. Some Bayesian approaches have leveraged Gaussian processes or Bayesian neural networks~\cite{kapoor2007active, roy2001toward, ucb} for uncertainty estimation. Gal et al.~\cite{dbal, gal2016dropout} demonstrated that Monte Carlo (MC) dropout can be used for approximate Bayesian inference for active learning on small datasets. However, as discussed in later works~\cite{sener2018coreset, kirsch2019batchbald}, Bayesian techniques often do not scale well to large datasets due to batch sampling. In the non-Bayesian setting, uncertainty heuristics such as distance from the decision boundary, highest entropy, and expected risk minimization have been widely investigated~\cite{brinker2003incorporating, tong2001support, active_erm}. Ensembles have shown to be successful in representing uncertainty~\cite{qbcwithoutcoreset, suggestiveannotation} and can outperform MC dropout approaches~\cite{Beluch_2018_CVPR}.

Diversity-based methods focus on sampling images that represent the diversity in a set of unlabeled images. Sener et al. \cite{sener2018coreset} proposed a core-set approach for sampling diverse points by solving the k-Center problem in the last-layer embedding of a model. They also produced a theoretical bound on core-set loss that scales with the number of classes, meaning that performance deteriorates as class size increases. Furthermore, with high-dimensional data, distance-based representations including core-set often suffer from a convergence of pairwise distances in high dimensions, known as the \textit{distance concentration phenomenon} \cite{franccois2008high}. Variational adversarial active learning (VAAL) was proposed by \cite{vaal} as a task-agnostic diversity-based algorithm that samples data points using a discriminator trained adversarially to discern labeled and unlabeled points on the latent space of a variational auto-encoder.

Hybrid approaches leverage both uncertainty and diversity. BatchBALD~\cite{kirsch2019batchbald} was proposed as a tractable remedy for non-jointly informative sampling in BALD. Using a Monte Carlo estimator, BatchBALD approximates the mutual information between a batch of points and the model's parameters and greedily generates a batch of samples. Rather than using model outputs as is standard for most uncertainty-based approaches, BADGE~\cite{ash2020deep} utilizes the \textit{gradient} of the predicted category with respect to the last layer of the model. These gradients provide an uncertainty measure as well as an embedding that can be clustered via $\text{k-means++}$~\cite{kmeans} initialization to select diverse points.

Although active learning research in semantic segmentation has been a bit more sparse, approaches largely parallel those of classification methods such as in the use of MC Dropout~\cite{gorriz2017costeffectivemelanoma}, feature entropy, and conditional geometric entropy for uncertainty measures~\cite{Konyushkova_2015_ICCV}. Diversity-based samplers like VAAL~\cite{vaal} were also shown to be effective on semantic segmentation. Recently, \cite{Casanova2020Reinforced} proposed a deep reinforcement learning approach for selecting regions of images to maximize Intersection over Union (IoU). 

\noindent \textbf{Entropy regularization}
Entropy regularization has been widely used in parametric models of posterior probabilities to benefit from unlabeled data or partially labeled data~\cite{grandvalet2006entropy}.  Entropy was also proposed to be used for clustering data and training a classifier simultaneously~\cite{krause2010discriminative}. In the field of domain adaptation, \cite{tzeng2015simultaneous} used cross-entropy between target activation and soft labels to exploit semantic relationships in label space whereas~\cite{springenberg2015unsupervised} trained a discriminative classifier using unlabeled data by maximizing the mutual information between inputs and target categories. In~\cite{mme}, they used entropy optimization for semi-supervised domain adaptation. Recently~\cite{minent} proposed a fully test-time adaptation method that performs entropy minimization to adapt to the test data at inference time.

\noindent \textbf{Adversarial learning} Adversarial learning has been used for different problems such as generative modeling models~\cite{gan}, object composition~\cite{compositional}, representation learning~\cite{makhzani2015adversarial}, domain adaptation~\cite{adda},  continual learning~\cite{acl}, and active learning~\cite{vaal}. The use of an adversarial network enables the model to train in a fully-differentiable manner by adjusting to solve the \textit{minimax} optimization problem~\cite{gan}. In this work, we perform the minimax between a feature encoder and a classifier using Shannon entropy~\cite{shannon}.

\section{Minimax Active Learning}
In this section, we introduce \ours formulation for image classification and semantic segmentation settings. Our training procedure is shown in Algorithm~\ref{alg:mal}.

We consider the problem of learning a label-efficient model by iteratively selecting the most \textit{informative} samples from a given \textit{split} to be labeled by an oracle or a human expert. We start with an initial small pool of labeled data denoted as  $\mathcal{L}=\{(x^i_L,y^i_L)\}_{i=1}^{N_L}$, and a large pool of unlabeled data denoted as $\mathcal{U}=\{x^i_U\}_{i=1}^{N_U}$ where our goal is to populate a fixed sampling \textit{budget}, $b$, using an acquisition strategy to query samples from the unlabeled pool ($x_U \sim X_U$) such that the expected loss is minimized. 

\subsection{\textbf{\ours} for image classification}
\ours starts with encoding data using a convolutional neural network (CNN), denoted as $F$, to a $d$-dimensional latent vector where features are normalized using an $\ell_2$ normalization. Similar to ~\cite{baseline++} in few-shot learning and~\cite{mme} in domain adaption, we exploit a cosine similarity-based classifier denoted as $C$ and parameterized by the weight matrix $\mathbf{W}\in\mathbb{R}^{d\times K}$, which takes the normalized features as input and maps them to $K$ class prototype vectors $[\textbf{w}_1, \textbf{w}_2, \cdots, \textbf{w}_K]$, where $K$ is the total number of classes in the dataset. The outputs of $C$ are converted to probability values $\mathbf{p}\in\mathbb{R}^K$ using a Softmax function ($\sigma$). %
We first train $F$ and $C$ by performing a $K$-way classification task on the already labeled data using a standard cross-entropy loss shown below 
\begin{equation}\label{eq:ce_loss}
\mathcal{L}_{\text{CE}} = -\mathbb{E}_{(x_L, y_L)\sim\mathcal{L}} \sum_{k=1}^{K} \mathbbm{1}_{[k=y_L]}  \log(\sigma(\frac{1}{T} \frac{\mathbf{W}^T F({x_L})}{||F({x_L})||}))
\end{equation} 
where the subscript $i=\{1,\cdots,N\}$ is dropped for simplicity. Eq. \ref{eq:ce_loss} learns the weight vectors $\mathbf{w}$ for each class by computing entropy which represents cosine similarity between each $\mathbf{w}$ and the output of the feature extractor. 

Our key idea is to learn a discriminative feature space by minimizing the intraclass distribution gap on the unlabeled data by adversarially maximizing the entropy so that all samples have equal distance to all prototypes. We formulate this minimax game using Shannon entropy~\cite{shannon}:
\begin{equation}\label{eq:minimax}
\mathcal{L}_{Ent} =\min_{F} \max_{C} \big(-\sum_{k=1}^{K} p(y=k|x_U)~\log p(y=c|x_U)\big)
\end{equation}
where we first minimize the entropy in $F$, and that is minimizing the cosine similarity score between unlabeled samples associated with the same prototype, which results in a more discriminative representation. Next, we maximize entropy in $C$, hence making a more uniform feature space. To achieve adversarial learning, the sign of gradients for entropy loss on unlabeled samples is flipped by a gradient reversal layer~\cite{grl}. 

The end goal of this minimax game is to generate a mixture of latent features associated with the labeled and unlabeled data, where a discriminator can be trained to map the latent representation to a binary label, which is $1$ if the sample belongs to $\mathcal{L}$ and is $0$ otherwise. We call this binary classifier $D$ and use a standard binary cross-entropy loss to train it.
\begin{equation}\label{eq:bce}
\begin{aligned}
\mathcal{L}_{BCE} = - \mathbb{E} [\log(\frac{1}{T} D(\frac{F(x_L)}{||F({x_L})||}))] &- \\ \mathbb{E} [\log(1- \frac{1}{T}D(\frac{F(x_U)}{||F(x_U)||}))]
\end{aligned}
\end{equation}

As shown in Algorithm \ref{alg:task}, at the end of each split, we train a model with the collected data and report our performance. We first initialize the \textit{task model}, denoted as $M$, using a pre-trained feature extractor $F$ from the \ours training process and use it to continually learn all the collected splits. 

\subsection{\textbf{\ours} for semantic segmentation}

For semantic segmentation, \ours similarly starts with encoding data using a convolutional neural network ($F$). However, the output is now a $d \times h \times w$ latent tensor where $h$ and $w$ are the height and width of the output filters, respectively as the segmentation task requires spatial information. After $\ell_2$ normalization, we pass this tensor through a classifier $C$, which uses $K$ prototype convolution filters $[\textbf{w}_1, \textbf{w}_2, \cdots, \textbf{w}_K]$ to produce a $K \times h \times w$ tensor. This tensor is then upsampled to the original image size and converted to probability tensor using the Softmax function, the result being $K \times H \times W$ where $H$ and $W$ are the height and width of the input images, respectively. $F$ and $C$ are trained using a pixel-average cross-entropy loss, where here the label is a $H \times W$ segmentation map.

To learn a discriminative feature space, we similarly train $F$ to minimize entropy and adversarially train $C$ to maximize it. However, in this formulation, entropy is taken to be the pixel-average in the output. To have our latent feature space contain \textit{labeledness} information, we again train a discriminator as in above, but our discriminator model now takes in $d \times h \times w$ tensors as input and outputs a binary prediction for whether or not a sample is labeled.

\subsection{Sampling strategy in \ours}
The sampling strategy in \ours is shown in Algorithm~\ref{alg:sampling}. The key idea to our approach is that we have two conditions to select a sample for labeling. (1) We use the probability associated with the discriminator's predictions as a score to rank samples based on their \textit{diversity}, which can be interpreted as how similar they are to previously seen data. The closer the probability is to zero, the more confident $D$ is that it comes from the unlabeled pool. (2) We use the entropy obtained by $C$ on the unlabeled data and use it as a score to rank samples based on how far they are from class prototypes. We take the top $b$ samples that meet both criteria. Unlike the uncertainty-based methods, our approach is not vulnerable to outliers as condition (1) prevents it because outliers never achieve a high confidence score from the discriminator because they are not similar to any samples that $D$ has seen. On the other hand, unlike diversity-based approaches, \ours is task-aware and can identify unlabeled samples far from the learned class prototypes.

\begin{algorithm}[tb]
	\caption{Minimax Active Learning (\ours)} \label{alg:mal}
	\begin{algorithmic}[1]
		\renewcommand{\algorithmicrequire}{\textbf{Input:}}
		\renewcommand{\algorithmicensure}{\textbf{Output:}}
		\REQUIRE Labeled pool $\mathcal{L}$, Unlabeled pool $\mathcal{U}$, Initialized models for ${\theta_F}$, $\theta_C$, $\theta_D$
		\REQUIRE Hyperparameters: epochs, $M$,  $\alpha_1$,  $\alpha_2$,  $\alpha_3$
		\FOR {$e = 1$ \text{to epochs}}
		\STATE sample $(x_L, y_L) \sim \mathcal{L}$
		\STATE sample $x_U \sim \mathcal{U}$
		\STATE Compute $\mathcal{L}_{\mathrm{CE}}$ by using Eq. \ref{eq:ce_loss} 
		\STATE $\theta'_{F} \gets \theta_{F} - \alpha_1 \nabla \mathcal{L}_\mathrm{CE} $
		\STATE $\theta'_{C} \gets \theta_{C} - \alpha_2 \nabla \mathcal{L}_\mathrm{CE} $
		\vspace{3pt}
		\STATE Compute $\mathcal{L}_{Ent}$ by using Eq. \ref{eq:minimax} 
		\vspace{3pt}     
		\STATE $\theta'_{F} \gets \theta_{F} + \alpha_1 \nabla \mathcal{L}_\mathrm{Ent} $
		\STATE $\theta'_{C} \gets \theta_{C} - \alpha_2 \nabla \mathcal{L}_\mathrm{Ent} $
		\STATE Compute $\mathcal{L}_\mathrm{D}$ by using Eq. \ref{eq:bce} 
		\vspace{1pt}
		\STATE $\theta'_{D} \gets \theta_{D} - \alpha_3 \nabla \mathcal{L}_\mathrm{D} $
		\vspace{3pt}
		\ENDFOR
		\OUTPUT Trained ${\theta_F}$, $\theta_C$, $\theta_D$
	\end{algorithmic} 
\end{algorithm}

\begin{algorithm}[t]
	\caption{Sampling Strategy in \ours} \label{alg:sampling}
	\begin{algorithmic}[1]
		\renewcommand{\algorithmicrequire}{\textbf{Input:}}
		\renewcommand{\algorithmicensure}{\textbf{Output:}}
		\REQUIRE $b, X_{L}, X_{U}$
		\ENSURE $X_{L}, X_{U}$
		\STATE Select samples ($X_s$) with $\min_b\{\theta_{D}(F(x))\}$ and $\max_b\{\theta_{C}(F(x))\}$%
		\STATE $Y_{o} \leftarrow \mathcal{ORACLE}(X_{s})$
		\STATE $(X_L,Y_L) \leftarrow (X_L,Y_L) \cup (X_s,Y_o)$
		\STATE $X_U \leftarrow X_U - {X_{s}}$
		\OUTPUT $X_{L}, X_{U}$
	\end{algorithmic} 
\end{algorithm}

\begin{algorithm}[t]
	\caption{Training for Main Task in \ours} \label{alg:task}
	\begin{algorithmic}[1]
		\renewcommand{\algorithmicrequire}{\textbf{Input:}}
		\renewcommand{\algorithmicensure}{\textbf{Output:}}
		\REQUIRE Labeled pool $\mathcal{L}$, Pre-trained ${F_\theta}$ from Alg. \ref{alg:mal}, $M_\theta$
		\REQUIRE Hyperparameters: epochs, $\alpha_4$
		\STATE Initialize $M_\theta$ with pre-trained $F$ backbone
		\FOR {$e = 1$ \text{to epochs}}
		\STATE Compute $\mathcal{L}_{\text{CE}}$ by using Eq. \ref{eq:ce_loss} 
		\STATE $\theta'_{M} \gets \theta_{M} - \alpha_4 \nabla \mathcal{L}_\mathrm{CE} $
		\vspace{3pt}
		\ENDFOR
		\OUTPUT $M_\theta$
	\end{algorithmic} 
\end{algorithm}

\section{Experiments}\label{sec:experiments}
In this section, we review the benchmark datasets and baselines used in our evaluation as well as the implementation details. 

\noindent \textbf{Datasets.} We have evaluated \ours on two common vision tasks: image classification and semantic segmentation. For large-scale image classification, we have used \textbf{\imagenet}~\cite{imagenet} with more than $1.2$M images of $1000$ classes and \textbf{\inat}~\cite{inat}, which is a heavily imbalanced dataset with more than $430$K natural images containing $8142$ species. We have also used \textbf{\cifar}~\cite{cifar} with $60$K images of size $32\times32$ as a popular benchmark in the field. To highlight the benefits of evaluating active learning algorithms in more realistic scenarios, we also created an imbalanced version of \cifar (see Section \ref{sec:classification}). For semantic segmentation, we evaluate our method on $\textbf{BDD100K}$ \cite{bdd100k} and $\textbf{Cityscapes}$~\cite{cityscapes} datasets both of which have $19$ classes. \bdd is a diverse driving video dataset with $7$K images with full-frame semantic segmentation annotations collected from distinct locations in the United State. $\text{Cityscapes}$ is also another driving video dataset containing $3475$ frames with segmentation annotations recorded in street scenes from different cities in Europe. We have resized images in 
\bdd and \city to $688\times688$. The statistics of these datasets are summarized in the appendix. 

\noindent \textbf{Baselines}: we compare the performance of \ours against several other diversity and uncertainty based active learning approaches including VAAL~\cite{vaal}, Coreset~\cite{sener2018coreset}, Entropy~\cite{wang2014label}, BALD~\cite{dbal}. We also show results for \textit{Random sampling} in which samples are uniformly sampled at random, without replacement, from the unlabeled data and serves as a competitive baseline in active learning. We also demonstrate how our semi-supervised learning strategy (adversarial entropy optimization) can replace the supervised learning regime in random sampling, uncertainty, or diversity-based active learning techniques. To make an appropriate and fair comparison, we integrated all the baselines into our code-base to ensure using identical training protocols.

\noindent \textbf{Performance measurement.}
We evaluate the performance of \ours on the main task at the end of labeling predefined splits of the dataset using the task model $M$. In classification we measure the class prediction accuracy using ResNet18~\cite{resnet} as our $M$ network whereas, in segmentation, we compute mean IoU on the collected data where $M$ is a dilated residual network (DRN)~\cite{drn}. Algorithm \ref{alg:task} shows how we train $M$ by initializing it with the feature extractor module ($F$) learned during \ours's training which results in \ours achieving a higher mean accuracy \textit{only by using the initial labeled pool and the unlabeled data before annotating any new instances}, compared to all the baselines in all the experiments (see Section \ref{sec:classification} and \ref{sec:segmentation}). Results for all our experiments are averaged over $3$ runs. %

\subsection{\textbf{\ours} on image classification benchmarks} \label{sec:classification} 

\noindent \textbf{Implementation details.} Images in \imagenet and \inat are resized to $256$ and then random-cropped to $224$. We used random horizontal flips for data augmentation for all the datasets. For \imagenet, we used $5\%$ of the entire dataset for the initial labeled pool and considered a budget size of $5\%$ for each split. These numbers are both $10\%$ for \inat and $2\%$ for \cifar dataset. The pool of unlabeled data contains the rest of the training set from which samples are selected to be annotated by the oracle. Once labeled, they will be added to the initial training set and training is repeated on the new training set. We assume the oracle is \textit{ideal} and provides accurate labels. The architecture used in the task module for image classification is ResNet18~\cite{resnet} pretrained with ImageNet except for the ImageNet experiment itself. When comparing to BALD we additionally introduce a $p=0.2$ dropout layer before the final average pooling during training. The discriminator is a $3$-layer multilayer perceptron (MLP) with ReLU non-linearities~\cite{relu} and Adam~\cite{kingma2014adam} is used as the optimizer for $F$, $C$, and $D$. %

\begin{table}[t]
\caption{Performance on \textbf{\imagenet}. Top-1 denotes the accuracy obtained with supervised learning on full dataset using ResNet18. Results are averaged over $3$ runs and standard deviations are shown in parenthesis.}
\label{tab:imagenet}
\vskip 0.1in
\begin{center}
\begin{footnotesize}
\begin{tabular}{lcccc}
\toprule
          & 5\% & 10\% & 15\% & 20\% \\
\midrule
Random    & 26.6(0.3) 	& 35.1(0.3) 	& 42.2(0.4) 	& 46.1(0.3)     \\
Entropy   & 26.6(0.3) 	& 36.1(0.2) 	& 43.4(0.3) 	& 46.2(0.1) 	\\
Coreset   & 26.6(0.3) 	& 37.4(0.2)	& 43.7(0.3) 	& 47.4(0.2) 	\\
BALD      & 26.6(0.3) 	& 37.0(0.5) 	& 44.6(0.3) 	& 48.1(0.4)     \\
VAAL      & 26.6(0.3) 	& 37.8(0.3) 	& 45.8(0.4) 	& 48.3(0.5) 	\\
\textbf{\ours (Ours)} & \textbf{32.7(0.3)} & \textbf{42.9(0.3)}	& \textbf{50.1(0.1)}	& \textbf{52.9(0.3)}\\
\hline
Top-1 & \multicolumn{4}{c}{61.5(0.8)} \\
\bottomrule
\end{tabular}
\end{footnotesize}
\end{center}
\vskip -0.1in
\end{table}

\begin{table}[t]
\caption{Performance on \textbf{\inat}. Top-1 denotes the accuracy obtained with supervised learning on full dataset using a ResNet18 pretrained on ImageNet. Results are averaged over $3$ runs and standard deviations are shown in parenthesis.}
\label{tab:inat}
\vskip 0.1in
\footnotesize
\begin{center}
\begin{tabular}{lcccc}
\toprule
 \% Labeled data     & 10\% & 20\% & 30\% & 40\% \\
\midrule
Random  & 10.5(0.5)  & 19.0(0.4)  & 24.1(0.3) & 30.1(0.4) \\
Coreset & 10.5(0.5)	& 18.0(0.3)  & 24.6(0.2) & 30.5(0.2) \\
VAAL    & 10.5(0.5)  & 19.5(0.1)  & 24.9(0.3) & 31.5(0.4)  \\
Entropy & 10.5(0.5)  & 20.3(0.3)  & 25.3(0.4) & 33.0(0.3) \\
BALD    & 10.5(0.5)	& 21.4(0.6)  & 26.8(0.5) & 34.9(0.4)  \\
\textbf{\ours(Ours)}           & \textbf{11.5(0.2)} 	& \textbf{29.2(0.4)} 	& \textbf{32.9(0.3)}	& \textbf{39.2(0.1)} \\
\hline
Top-1 & \multicolumn{4}{c}{46.7(0.2)} \\
\bottomrule
\end{tabular}
\end{center}
\vskip -0.1in
\end{table}

 \begin{table}[t]
\caption{Performance on \textbf{\cifar}. Top-1 denotes the accuracy obtained with supervised learning on full dataset using a ResNet18 pretrained on ImageNet. Results are averaged over $3$ runs and standard deviations are shown in parenthesis.}
\label{tab:cifar}
\vskip 0.1in
\begin{center}
\scriptsize
\begin{tabular}{lccccc}
\toprule
          & 2\% & 4\% & 6\% & 8\% & 10\% \\
\midrule
Entropy   &  24.2(0.5)   & 30.9(0.3) 	& 36.0(0.3)     & 38.4(0.4)   & 40.2(0.6) \\
Coreset   &  24.2(0.5) 	& 31.8(0.3) 	& 36.8(0.5) 	   & 39.7(0.3) 	& 40.4(0.5) 	\\
Random    &  24.2(0.5)	& 31.5(0.3) 	& 37.6(0.3) 	   & 40.0(0.2) 	& 40.4(0.4) 	\\
BALD      &  24.2(0.5) 	& 30.9(0.4)  & 40.1(0.2)     & 40.5(0.2)   & 41.1(0.5)	\\
VAAL      &  24.2(0.5) 	& 32.4(0.4) 	& 38.5(0.3) 	   & 40.9(0.3) 	& 41.6(0.4)	\\
\textbf{\ours (Ours)} & \textbf{26.8(0.3)} 	& \textbf{35.2(0.4)} & \textbf{40.9(0.3)}	& \textbf{43.4(0.2)}  &  \textbf{46.1(0.2)}  \\ %
\hline
Top-1 & \multicolumn{4}{c}{60.7(0.1)} \\
\bottomrule
\end{tabular}
\end{center}
\vskip -0.1in
\end{table}

\begin{table}[t]
\caption{Performance on \textbf{\imbalancedcifar}. Top-1 denotes the accuracy obtained with supervised learning on full \imbalancedcifar dataset with $14586$ images using a ResNet18 pretrained on ImageNet. Results are averaged over $3$ runs and standard deviations are shown in parenthesis.}
\label{tab:imbalanced}
\vskip 0.1in
\begin{center}
\footnotesize
\begin{tabular}{lcccc}
\toprule
  \# Labeled data       & 1458 &  2916 & 4375 & 5834 \\
\midrule
Random & 15.5(0.4) & 21.0(0.3) & 23.8(0.3) & 26.3(0.3) \\  
BALD & 15.5(0.4) & 21.5(0.5) & 23.9(0.7) & 27.3(0.6)  \\  %
Coreset & 15.5(0.4) & 21.4(0.7) & 24.9(0.4) & 27.7(0.3)  \\
VAAL &  15.5(0.4)   & 21.8(0.6) & 25.0(0.6) & 27.7(0.5) \\
Entropy & 15.5(0.4) & 22.3(0.3) & 25.6(0.5) & 27.8(0.4) \\
\textbf{\ours (Ours)} & \textbf{20.4(0.5)} 	& \textbf{25.4(0.5)} & \textbf{28.4(0.4)}	  &  \textbf{30.2(0.3)}  \\ %
\hline
Top-1 & \multicolumn{4}{c}{34.2(0.8)} \\
\bottomrule
\end{tabular}
\end{center}
\vskip -0.1in
\end{table}

\noindent \textbf{\ours performance on \imagenet.} Table~\ref{tab:imagenet} shows mean accuracy obtained by our method compared to prior works and Random sampling for $4$ splits from $5\%-20\%$ data labeling. Top-1 denotes the accuracy obtained with supervised learning on the full dataset using a randomly initialized ResNet18 architecture. We first want to highlight that, in \textit{all} the experiments, \ours achieves higher mean accuracy in the beginning \textit{before} selecting instances for labeling due to the effective semi-supervised learning strategy performed during adversarial training of $F$ and $C$. We improve the state-of-the-art by more than $100\%$ increase in the gap between the accuracy achieved by the previous state-of-the-art method (VAAL) and Random sampling. As can be seen in Table~ \ref{tab:imagenet}, this improvement can be also viewed in the number of samples required to achieve a specific accuracy. For instance, the accuracy of $42.9\pm0.8\%$ is achieved by \ours using $10\%$ of the data ($128$K images) whereas VAAL should be provided with almost $32$K more images and Coreset and Random sampling with $64$K more labeled images to obtain the same performance. \ours can achieve $52.9\%$ accuracy using only $20\%$ of the data while the accuracy on the full dataset is $61.5\pm0.8$.

\noindent \textbf{\ours performance on \inat.} Table~\ref{tab:inat} shows the accuracy achieved on \inat as a large-scale heavily long-tailed distribution with $8142$ classes. \ours achieves mean accuracy of 39.2\% using 40\% of the data while supervised training on the entire dataset with our setup yields 46.7\%. From the labeling-reduction point of view, while \ours achieves 29.2\% by selecting 20\% of the data, VAAL and Random sampling require nearly 10\% and 20\% more annotations to obtain the same accuracy, respectively. This experiment highlights the value of evaluating on realistic datasets representing real-world difficulties associated with active learning. %

\noindent \textbf{\ours performance on \cifar.} Table~\ref{tab:cifar} shows the performance of \ours on \cifar which is a common benchmark used in active learning literature. \ours achieves mean accuracy of $46.1\%$ by using only $10\%$ of the labels whereas using the entire
dataset yields accuracy of $60.7\%$. The most competitive baseline, VAAL, obtains $41.6\%$ while all other baselines including Random, Coreset, Entropy, and BALD reach $~40\%$. \ours consistently requires $2\%-3\%$ less data which is equivalent to $1000$-$1500$ fewer number of labels compared to Random/Coreset/Entropy/BALD methods in order to achieve the same accuracy. This number is $1\%$ for the most competitive baseline, VAAL.%

\noindent \textbf{\ours performance on \imbalancedcifar.} Prior work has largely considered using datasets with equally distributed images among classes~\cite{vaal,sener2018coreset,qbcwithoutcoreset,wang2014label}. Our results on \inat dataset show that the majority of the algorithms in the literature fail in such scenarios, and hence, only evaluating on balanced datasets cannot accurately capture the performance of active learning algorithms. In an effort to mitigate this problem, we created an imbalanced version of \cifar by keeping random \textit{ratios} from each class, such that the least populated class has at least five instances while allowing for imbalances up to $100\text{x}$. We use imbalance ratios of $\{10^{-2}, 10^{-1.5}, 10^{-1}, 10^{-0.5}, 10^0\}$. We use each ratio for $20$ randomly chosen classes without replacement. 

Table~\ref{tab:imbalanced} shows the results for the \imbalancedcifar experiment. \ours outperforms the baselines across all the splits. While the highest accuracy obtained by supervised training on all the 14586 images is 34.2\%, \ours achieves 30.2\% by labeling 5834 images only. %

\subsection{\textbf{\ours} on image segmentation benchmarks}\label{sec:segmentation}

\noindent\textbf{Implementation details.} Similar to the image classification setup, we used random horizontal flips for data augmentation. Images and segmentation labels are resized to $688 \times 688$, with nearest-neighbor interpolation being used on the segmentation map. On both \bdd and \city datasets, for the initial labeled pool, we used $10\%$ of the entire dataset, and for active learning, we used a budget size of $5\%$ of the full training set. %
The architecture used in $F$ is DRN~\cite{drn}, and $C$ is a convolutional layer followed by an upsampling layer. Our setup for $C$ is similar to what is used in the full segmentation model in the DRN paper. For our discriminator, instead of using a $3$-layer MLP, we use a $3$-layer CNN followed by a small fully-connected layer. Finally, for our choice of the optimizer, we use Adam~\cite{adam}. 

\begin{table}[t]
\caption{Performance on \textbf{\city}. Top-1 denotes the \%mIoU obtained with supervised learning on full dataset using a DRN architecture pretrained on ImageNet. Results are averaged over $3$ runs and standard deviations are shown in parenthesis.}
\label{tab:city}
\vskip 0.1in
\begin{center}
\scriptsize
\begin{tabular}{lccccc}
\toprule
\% Labeled data       & 10\% & 15\% & 20\% & 25\% & 30\%  \\ %
\midrule
Random   & 46.2(0.8) & 49.4(1.8) & 49.1(1.4) & 52.7(0.9) &  53.4(0.7)     \\    %
Entropy  & 46.2(0.8)  & 48.1(1.4) & 50.4(0.9) & 52.1(0.9) & 53.9(0.5) \\
Coreset  & 46.2(0.8) & 48.9(1.8) & 51.6(0.8) & 53.2(1.0) & 54.7(0.8)  \\ %
BALD     &  46.2(0.8) & 49.1(0.9) & 51.8(1.1)  & 53.4(1.1) & 54.6(0.9)   \\ %
VAAL 	 &  46.2(0.8) &  49.7(0.9) & 52.3(1.4) & 54.1(0.9) & 55.5(0.7)    \\	%
\textbf{\ours (Ours)} & \textbf{48.9(0.7)} 	& \textbf{51.5(0.8)} & \textbf{55.7(0.8)}	& \textbf{57.1(0.6)}  &   \textbf{58.4(0.4)} \\ %
\hline
Top-1 & \multicolumn{5}{c}{61.9(0.7)} \\
\bottomrule
\end{tabular}
\end{center}
\vskip -0.1in
\end{table}

\begin{table}[t]
\caption{Performance on \textbf{\bdd}. Top-1 denotes the \%mIoU obtained with supervised learning on full dataset using a DRN architecture pretrained on ImageNet. Results are averaged over $3$ runs and standard deviations are shown in parenthesis.}
\label{tab:bdd}
\vskip 0.1in
\begin{center}
\scriptsize
\begin{tabular}{lccccc}
\toprule
          & 10\% & 15\% & 20\% & 25\% & 30\%  \\
\midrule
Random  & 35.1(1.7) & 36.5(1.2) & 37.6(0.8) & 38.9(0.9) & 40.0(0.9) \\
Coreset & 35.1(1.7) & 36.8(1.8) & 38.3(1.6) & 39.5(0.8) &  40.5(0.9) \\
Entropy & 35.1(1.7) & 36.5(0.9) &  38.6(1.2)   & 39.6(0.6)  & 41.4(0.8)  \\
VAAL & 35.1(1.7) & 37.4(1.2) & 39.2(1.4) & 39.7(0.7) & 41.5(0.5) \\
\textbf{\ours (Ours)} & \textbf{39.3(1.3)} 	& \textbf{41.4(1.6)} & \textbf{44.1(0.9)}	& \textbf{45.3(0.9)}  &  \textbf{45.9(0.9)}  \\
\hline
Top-1 & \multicolumn{5}{c}{51.2(1.1)} \\
\bottomrule
\end{tabular}
\end{center}
\vskip -0.1in
\end{table}

\noindent \textbf{\ours performance on \city.} Table~\ref{tab:city} demonstrates our results on \city dataset. \ours consistently demonstrate better performance by achieving the highest mean IoU across different labeled data ratios. \ours is able to achieve $\%$mIoU of 58.4\% using only 30\% labeled data (889 fully segmented images) on \city whereas training with all the data yields 61.9\%. On \bdd dataset, \ours achieves $\sim$4\% better mIoU across all the splits.  In terms of required labels by each method, on \city, \ours needs 20\% of the annotations to reach 55\% of mIoU whereas the best baseline, VAAL, demands nearly 10\% more labeled images to obtain the same mIoU. This gap is larger on \bdd such that VAAL requires twice more annotations (30\%) than \ours (15\%) to achieve nearly 40\%mIoU. Considering the difficulties in full-frame instance segmentation, \ours is able to effectively reduce the required time and effort for such dense annotations.

\section{Ablation study}
In this section, we take a deeper look into our model by performing an ablation study on \imagenet and \bdd to inspect the contribution of the key modules in \ours as well as its sampling strategy. We compare our results with the most complete version of \ours and Random sampling. As shown by our ablations, each aspect of \ours is essential to its overall performance, indicating that the individual components operate in a complementary fashion to benefit the overall method. %

Table~\ref{tab:ablation-imagenet} and Table~\ref{tab:ablation-bdd} illustrate our ablation study on \imagenet and \bdd datasets, respectively. For the first ablation, we eliminate the role of minimax entropy between $F$ and $C$ such that they only perform regular classification on the labeled data and no adversarial loss is computed on the unlabeled data using entropy. In other words, we ablate the semi-supervised learning aspect of \ours and hence convert it to a Supervised active Learning (SL) strategy with a random (SL-Random), uncertainty (SL-Unc), or diversity-based (SL-Div) sampler. Note that for SL-Div, we still train $D$ on $F$'s feature space to predict the labeledness. For SL-Unc we eliminate $D$ and keep $F$ and $C$ to compute entropy values to select samples for labeling. As can be seen in for both datasets in Table~\ref{tab:ablation-imagenet} and Table~\ref{tab:ablation-bdd} , SL-Random and regular Random sampling perform on par while SL-Div and SL-Unc settings perform worse than them. %

We refer to \ours in Tables \ref{tab:ablation-imagenet} and \ref{tab:ablation-bdd} by SSL-Div-Unc to distinguish it from its ablated versions, SSL-Random, SSL-Div, and SSL-Unc, which are shown to highlight the effect of the sampling criteria in \ours. We observed that using \ours with entropy as the selection criterion (SSL-Unc) is outperformed by SSL-Random and SSL-Div by a large margin early in the process. On the other hand, using \ours with random sampling only yields sub-optimal results in the beginning. Among SSL-Unc and SSL-Div, diversity-based sampling plays a more significant role in obtaining better performance. Overall, \ours, performs best with its hybrid sampling strategy in which it benefits from diversity and uncertainty-based samplings together.

 \begin{table}[t]
\caption{Ablation results on analyzing \ours variants on \imagenet. We showed the effect of semi-supervised versus supervised active learning strategies as well as the sampling criteria used in \ours. Results are averaged over $3$ runs and standard deviations are shown in parenthesis.}
\label{tab:ablation-imagenet}
\vskip 0.1in
\begin{center}
\scriptsize
\begin{tabular}{lccccc}
\toprule
          & 5\% & 10\% & 15\% & 20\% \\
\midrule
SL-Unc      &  26.9(0.3) 	& 35.2(0.4) 	& 40.2(0.1) 	   & 44.2(0.5) 	\\
SL-Div      &  26.8(0.2) 	& 35.6(0.3) 	& 41.1(0.4) 	   & 44.3(0.2) 	\\ 
SL-Random   &  26.9(0.3)     & 35.4(0.2) 	& 42.3(0.6)     & 45.7(0.4)   \\ \midrule
Random      &  26.6(0.3) 	& 35.1(0.3) 	& 42.2(0.4) 	   & 46.1(0.3)   \\ \midrule
SSL-Unc	    &  27.6(0.4)	    & 36.2(0.1) 	& 46.9(0.7)	   & 49.5(0.5) 	\\ 
SSL-Random  &  27.6(0.4)	    & 39.1(0.3) 	& 47.2(0.6) 	   & 49.8(0.2) 	\\
SSL-Div     &  27.6(0.4) 	& 38.8(0.3)  & 48.1(0.4)     & 50.0(0.2)   \\
\textbf{SSL-Div-Unc (\ours-ours)} & \textbf{32.7(0.3)} & \textbf{42.9(0.3)}	& \textbf{50.1(0.1)}	& \textbf{52.9(0.3)}\\
\hline
Top-1 & \multicolumn{4}{c}{61.5(0.8)} \\
\bottomrule
\end{tabular}
\end{center}
\end{table}

It is worth mentioning that the high-level outcome of our ablation study is aligned with those found in~\cite{vaal}. They found using unlabeled data (unsupervised reconstruction of images in their case) had the largest effect on their overall performance. Similarly, we found the same for semi-supervised learning in \ours. However, we showed unlabeled data can be utilized more effectively when they are used in an adversarial game with labeled data in a semi-supervised learning fashion.

 \begin{table}[t]
\caption{Ablation results on analyzing \ours variants on \bdd. We showed the effect of semi-supervised versus supervised active learning strategies as well as the sampling criteria used in \ours. Results are averaged over $3$ runs and standard deviations are shown in parenthesis.}
\label{tab:ablation-bdd}
\vskip 0.1in
\begin{center}
\scriptsize
\begin{tabular}{lcccc}
\toprule
          & 10\% & 15\% & 20\% & 25\%  \\
\midrule
SL-Unc      &  34.2(0.9) 	& 35.2(0.4) 	& 36.4(1.1) 	   & 37.4(1.4)   \\
SL-Div      &  34.5(1.5) 	& 35.6(0.3) 	& 36.9(1.3) 	   & 37.8(1.5) 	 \\ 
SL-Random   &  34.2(0.9)     & 35.4(0.2) 	& 37.4(1.0)     & 38.0(1.0)   \\ \midrule
Random      &  35.1(1.7)     & 36.5(1.2)  & 37.6(0.8)     & 38.9(0.9)    \\ \midrule
SSL-Unc	    &  37.3(1.2)	    & 38.2(1.2) 	& 41.9(0.7)	   & 42.8(0.9) 	\\ 
SSL-Random  &  37.3(1.2)	    & 38.1(0.8) 	& 42.1(0.8) 	   & 42.4(0.8) 	\\
SSL-Div     &  37.3(1.2) 	& 38.4(0.9)  & 42.8(0.5)     & 43.1(1.0)    \\
\textbf{SSL-Div-Unc (\ours-ours)} & \textbf{39.3(1.3)} 	& \textbf{41.4(1.6)} & \textbf{44.1(0.9)}	& \textbf{45.3(0.9)}   \\
\hline
Top-1 & \multicolumn{4}{c}{51.2(1.1)} \\
\bottomrule
\end{tabular}
\end{center}
\vskip -0.1in
\end{table}

\subsection{Choice of the network architecture for $F$:}\label{sec:arch}
In order to assure \ours is insensitive to the ResNet18 architecture used in our classification experiments, we also used VGG~\cite{vgg} network architecture in \ours and VAAL~\cite{vaal}. Table \ref{tab:arch} shows the performance of our method is robust to the choice of the architecture by having consistently better performance over VAAL on \cifar using VGG16-BN and ResNet18. Task learner and F share the same CNN. For the classifier head in the task learner using VGG backbone, we used only one fully connected layer after the convolutional layers with $(512\times7\times7) \times 100$ where $(512\times7\times7)$ is the output size of the CNN and 100 is total number of classes in \cifar. This results in having 14.72\text{M}, 12.90\text{M} and 13.11\text{M}  parameters in F, C, and D, respectively while T has 17.23\text{M} parameters. In our ResNet18 model, F, C, D, and T have 11.18\text{M}, 156.93\text{K}, 525.83\text{K}, and 11.23\text{M} parameters, respectively.

 \begin{table}[t]
\caption{Performance of \ours using ResNet18 and VGG16 on CIFAR100 versus VAAL on the same architecture.}
\label{tab:arch}
\vskip -0.1in
\begin{center}
\scriptsize
\begin{tabular}{lcccc}
\toprule
          & 2\% & 4\% & 6\% & 8\%  \\
\midrule
VAAL-VGG16      &   25.7(0.9)    & 35.7(0.4)       &  44.2(1.3)      &  47.9(0.8)  \\  
\textbf{\ours-VGG16}  	& \textbf{29.8(0.4)}  & \textbf{39.3(0.3)} & \textbf{45.5(0.2)}	&  \textbf{49.5(0.3)}   \\ \hline
Top-1 VGG-16 & \multicolumn{4}{c}{71.9(0.1)} \\
\midrule 
VAAL-ResNet18     &  24.2(0.5) 	& 32.4(0.4) 	& 38.5(0.3) 	& 40.9(0.3) 	\\
\textbf{\ours-ResNet18} & \textbf{26.8(0.3)} 	& \textbf{35.2(0.4)} & \textbf{40.9(0.3)}	& \textbf{43.4(0.2)} \\
\midrule
Top-1 ResNet18 & \multicolumn{4}{c}{60.7(0.1)} \\
\bottomrule
\end{tabular}
\end{center}
\vskip -0.1in
\end{table}

\section{Conclusion}

In this paper, we proposed a pool-based semi-supervised active learning algorithm, \ours, that learns a discriminative representation on the unlabeled data in an adversarial game between a feature extractor and a cosine similarity-based classifier while minimizing the expected loss on the labeled data. We train a binary classifier on this mixture of latent features to distinguish between the labeled and unlabeled pool and implicitly learn the uncertainty for the samples deemed to be from the unlabeled pool. We introduced a hybrid sampling strategy that selects samples that are (i) most different from the already labeled data and (ii) farthest from class prototypes learned by a cosine-similarity classifier. We demonstrate excellent improvement over the existing methods on small and large-scale image classification and semantic segmentation driving datasets.

{\small
\bibliographystyle{ieee_fullname}
\bibliography{egbib}

\begin{thebibliography}{10}\itemsep=-1pt

\bibitem{kmeans}
David Arthur and Sergei Vassilvitskii.
\newblock K-means++: The advantages of careful seeding.
\newblock In {\em Proceedings of the Eighteenth Annual ACM-SIAM Symposium on
  Discrete Algorithms}, SODA '07, page 1027–1035, USA, 2007. Society for
  Industrial and Applied Mathematics.

\bibitem{ash2020deep}
Jordan~T. Ash, Chicheng Zhang, Akshay Krishnamurthy, John Langford, and Alekh
  Agarwal.
\newblock Deep batch active learning by diverse, uncertain gradient lower
  bounds.
\newblock In {\em Eighth International Conference on Learning Representations
  (ICLR)}, April 2020.

\bibitem{compositional}
Samaneh Azadi, Deepak Pathak, Sayna Ebrahimi, and Trevor Darrell.
\newblock Compositional gan: Learning image-conditional binary composition.
\newblock {\em arXiv preprint arXiv:1807.07560}, 2018.

\bibitem{beluch2018ensemble}
William~H Beluch, Tim Genewein, Andreas N{\"u}rnberger, and Jan~M K{\"o}hler.
\newblock The power of ensembles for active learning in image classification.
\newblock In {\em Proceedings of the IEEE Conference on Computer Vision and
  Pattern Recognition}, pages 9368--9377, 2018.

\bibitem{Beluch_2018_CVPR}
William~H. Beluch, Tim Genewein, Andreas Nürnberger, and Jan~M. Köhler.
\newblock The power of ensembles for active learning in image classification.
\newblock In {\em Proceedings of the IEEE Conference on Computer Vision and
  Pattern Recognition (CVPR)}, June 2018.

\bibitem{brinker2003incorporating}
Klaus Brinker.
\newblock Incorporating diversity in active learning with support vector
  machines.
\newblock In {\em Proceedings of the 20th international conference on machine
  learning (ICML-03)}, pages 59--66, 2003.

\bibitem{Casanova2020Reinforced}
Arantxa Casanova, Pedro~O. Pinheiro, Negar Rostamzadeh, and Christopher~J. Pal.
\newblock Reinforced active learning for image segmentation.
\newblock In {\em International Conference on Learning Representations}, 2020.

\bibitem{baseline++}
Wei-Yu Chen, Yen-Cheng Liu, Zsolt Kira, Yu-Chiang~Frank Wang, and Jia-Bin
  Huang.
\newblock A closer look at few-shot classification.
\newblock In {\em International Conference on Learning Representations}, 2019.

\bibitem{cityscapes}
Marius Cordts, Mohamed Omran, Sebastian Ramos, Timo Rehfeld, Markus Enzweiler,
  Rodrigo Benenson, Uwe Franke, Stefan Roth, and Bernt Schiele.
\newblock The cityscapes dataset for semantic urban scene understanding.
\newblock In {\em Proceedings of the IEEE conference on computer vision and
  pattern recognition}, pages 3213--3223, 2016.

\bibitem{imagenet}
Jia Deng, Wei Dong, Richard Socher, Lie-Jia Li, Kai Li, and Li Fei-Fei.
\newblock {ImageNet: A Large-Scale Hierarchical Image Database}.
\newblock In {\em CVPR09}, 2009.

\bibitem{djolonga2020robustness}
Josip Djolonga, Jessica Yung, Michael Tschannen, Rob Romijnders, Lucas Beyer,
  Alexander Kolesnikov, Joan Puigcerver, Matthias Minderer, Alexander D'Amour,
  Dan Moldovan, et~al.
\newblock On robustness and transferability of convolutional neural networks.
\newblock {\em arXiv preprint arXiv:2007.08558}, 2020.

\bibitem{dosovitskiy2020image}
Alexey Dosovitskiy, Lucas Beyer, Alexander Kolesnikov, Dirk Weissenborn,
  Xiaohua Zhai, Thomas Unterthiner, Mostafa Dehghani, Matthias Minderer, Georg
  Heigold, Sylvain Gelly, et~al.
\newblock An image is worth 16x16 words: Transformers for image recognition at
  scale.
\newblock {\em arXiv preprint arXiv:2010.11929}, 2020.

\bibitem{ucb}
Sayna Ebrahimi, Mohamed Elhoseiny, Trevor Darrell, and Marcus Rohrbach.
\newblock Uncertainty-guided continual learning with bayesian neural networks.
\newblock In {\em International Conference on Learning Representations}, 2020.

\bibitem{acl}
Sayna Ebrahimi, Franziska Meier, Roberto Calandra, Trevor Darrell, and Marcus
  Rohrbach.
\newblock Adversarial continual learning.
\newblock {\em arXiv preprint arXiv:2003.09553}, 2020.

\bibitem{feng2019computer}
Xin Feng, Youni Jiang, Xuejiao Yang, Ming Du, and Xin Li.
\newblock Computer vision algorithms and hardware implementations: A survey.
\newblock {\em Integration}, 69:309--320, 2019.

\bibitem{franccois2008high}
Damien Fran{\c{c}}ois.
\newblock High-dimensional data analysis.
\newblock In {\em From Optimal Metric to Feature Selection}, pages 54--55. VDM
  Verlag Saarbrucken, Germany, 2008.

\bibitem{freund1997selective}
Yoav Freund, H~Sebastian Seung, Eli Shamir, and Naftali Tishby.
\newblock Selective sampling using the query by committee algorithm.
\newblock {\em Machine learning}, 28(2-3):133--168, 1997.

\bibitem{gal2016dropout}
Yarin Gal and Zoubin Ghahramani.
\newblock Dropout as a bayesian approximation: Representing model uncertainty
  in deep learning.
\newblock In {\em international conference on machine learning}, pages
  1050--1059, 2016.

\bibitem{dbal}
Yarin Gal, Riashat Islam, and Zoubin Ghahramani.
\newblock Deep bayesian active learning with image data.
\newblock {\em arXiv preprint arXiv:1703.02910}, 2017.

\bibitem{grl}
Yaroslav Ganin and Victor Lempitsky.
\newblock Unsupervised domain adaptation by backpropagation.
\newblock In {\em International conference on machine learning}, pages
  1180--1189. PMLR, 2015.

\bibitem{gilad2006query}
Ran Gilad-Bachrach, Amir Navot, and Naftali Tishby.
\newblock Query by committee made real.
\newblock In {\em Advances in neural information processing systems}, pages
  443--450, 2006.

\bibitem{gan}
Ian Goodfellow, Jean Pouget-Abadie, Mehdi Mirza, Bing Xu, David Warde-Farley,
  Sherjil Ozair, Aaron Courville, and Yoshua Bengio.
\newblock Generative adversarial nets.
\newblock In {\em Advances in neural information processing systems}, pages
  2672--2680, 2014.

\bibitem{gorriz2017costeffectivemelanoma}
Marc Gorriz, Axel Carlier, Emmanuel Faure, and Xavier Giro-i Nieto.
\newblock Cost-effective active learning for melanoma segmentation.
\newblock {\em arXiv preprint arXiv:1711.09168}, 2017.

\bibitem{grandvalet2006entropy}
Yves Grandvalet and Yoshua Bengio.
\newblock Entropy regularization., 2006.

\bibitem{minimax}
Marian Grend{\'a}r and Mari{\'a}n Grend{\'a}r.
\newblock Minimax entropy and maximum likelihood.
\newblock {\em arXiv preprint math.PR/0009129}, 2000.

\bibitem{resnet}
Kaiming He, Xiangyu Zhang, Shaoqing Ren, and Jian Sun.
\newblock Deep residual learning for image recognition.
\newblock In {\em Proceedings of the IEEE conference on computer vision and
  pattern recognition}, pages 770--778, 2016.

\bibitem{jaynes}
Edwin~T Jaynes.
\newblock Information theory and statistical mechanics. ii.
\newblock {\em Physical review}, 108(2):171, 1957.

\bibitem{kapoor2007active}
Ashish Kapoor, Kristen Grauman, Raquel Urtasun, and Trevor Darrell.
\newblock Active learning with gaussian processes for object categorization.
\newblock In {\em 2007 IEEE 11th International Conference on Computer Vision},
  pages 1--8. IEEE, 2007.

\bibitem{adam}
Diederik~P Kingma and Jimmy Ba.
\newblock Adam: A method for stochastic optimization.
\newblock {\em arXiv preprint arXiv:1412.6980}, 2014.

\bibitem{kingma2014adam}
Diederik~P Kingma and Jimmy Ba.
\newblock Adam: A method for stochastic optimization.
\newblock In {\em International Conference on Learning Representations}, 2015.

\bibitem{kirsch2019batchbald}
Andreas Kirsch, Joost van Amersfoort, and Yarin Gal.
\newblock Batchbald: Efficient and diverse batch acquisition for deep bayesian
  active learning.
\newblock In H. Wallach, H. Larochelle, A. Beygelzimer, F. d\textquotesingle
  Alch\'{e}-Buc, E. Fox, and R. Garnett, editors, {\em Advances in Neural
  Information Processing Systems 32}, pages 7026--7037. 2019.

\bibitem{kolesnikov2019big}
Alexander Kolesnikov, Lucas Beyer, Xiaohua Zhai, Joan Puigcerver, Jessica Yung,
  Sylvain Gelly, and Neil Houlsby.
\newblock Big transfer (bit): General visual representation learning.
\newblock {\em arXiv preprint arXiv:1912.11370}, 6, 2019.

\bibitem{Konyushkova_2015_ICCV}
Ksenia Konyushkova, Raphael Sznitman, and Pascal Fua.
\newblock Introducing geometry in active learning for image segmentation.
\newblock In {\em Proceedings of the IEEE International Conference on Computer
  Vision (ICCV)}, December 2015.

\bibitem{krause2010discriminative}
Andreas Krause, Pietro Perona, and Ryan Gomes.
\newblock Discriminative clustering by regularized information maximization.
\newblock {\em Advances in neural information processing systems}, 23:775--783,
  2010.

\bibitem{cifar}
Alex Krizhevsky and Geoffrey Hinton.
\newblock Learning multiple layers of features from tiny images.
\newblock Technical report, Citeseer, 2009.

\bibitem{kullback}
Solomon Kullback.
\newblock {\em Information theory and statistics}.
\newblock Courier Corporation, 1997.

\bibitem{qbcwithoutcoreset}
Weicheng Kuo, Christian H{\"a}ne, Esther Yuh, Pratik Mukherjee, and Jitendra
  Malik.
\newblock Cost-sensitive active learning for intracranial hemorrhage detection.
\newblock In {\em International Conference on Medical Image Computing and
  Computer-Assisted Intervention}, pages 715--723. Springer, 2018.

\bibitem{li2013adaptive}
Xin Li and Yuhong Guo.
\newblock Adaptive active learning for image classification.
\newblock In {\em Proceedings of the IEEE Conference on Computer Vision and
  Pattern Recognition}, pages 859--866, 2013.

\bibitem{mackay1992information}
David~JC MacKay.
\newblock Information-based objective functions for active data selection.
\newblock {\em Neural computation}, 4(4):590--604, 1992.

\bibitem{mahapatra2018efficient}
Dwarikanath Mahapatra, Behzad Bozorgtabar, Jean-Philippe Thiran, and Mauricio
  Reyes.
\newblock Efficient active learning for image classification and segmentation
  using a sample selection and conditional generative adversarial network.
\newblock In {\em International Conference on Medical Image Computing and
  Computer-Assisted Intervention}, pages 580--588. Springer, 2018.

\bibitem{makhzani2015adversarial}
Alireza Makhzani, Jonathon Shlens, Navdeep Jaitly, Ian Goodfellow, and Brendan
  Frey.
\newblock Adversarial autoencoders.
\newblock {\em arXiv preprint arXiv:1511.05644}, 2015.

\bibitem{adversarialsamplingactivelearning}
Christoph Mayer and Radu Timofte.
\newblock Adversarial sampling for active learning.
\newblock {\em arXiv preprint arXiv:1808.06671}, 2018.

\bibitem{mccallumzy1998employing}
Andrew~Kachites McCallumzy and Kamal Nigamy.
\newblock Employing em and pool-based active learning for text classification.
\newblock In {\em Proc. International Conference on Machine Learning (ICML)},
  pages 359--367. Citeseer, 1998.

\bibitem{metriclearning}
Thomas Mensink, Jakob Verbeek, Florent Perronnin, and Gabriela Csurka.
\newblock Metric learning for large scale image classification: Generalizing to
  new classes at near-zero cost.
\newblock In {\em European Conference on Computer Vision}, pages 488--501.
  Springer, 2012.

\bibitem{relu}
Vinod Nair and Geoffrey~E Hinton.
\newblock Rectified linear units improve restricted boltzmann machines.
\newblock In {\em ICML}, 2010.

\bibitem{preclustering}
Hieu~T Nguyen and Arnold Smeulders.
\newblock Active learning using pre-clustering.
\newblock In {\em Proceedings of the twenty-first international conference on
  Machine learning}, page~79. ACM, 2004.

\bibitem{lowshot}
Hang Qi, Matthew Brown, and David~G Lowe.
\newblock Low-shot learning with imprinted weights.
\newblock In {\em Proceedings of the IEEE conference on computer vision and
  pattern recognition}, pages 5822--5830, 2018.

\bibitem{roy2001toward}
Nicholas Roy and Andrew McCallum.
\newblock Toward optimal active learning through monte carlo estimation of
  error reduction.
\newblock {\em ICML, Williamstown}, pages 441--448, 2001.

\bibitem{mme}
Kuniaki Saito, Donghyun Kim, Stan Sclaroff, Trevor Darrell, and Kate Saenko.
\newblock Semi-supervised domain adaptation via minimax entropy.
\newblock In {\em Proceedings of the IEEE International Conference on Computer
  Vision}, pages 8050--8058, 2019.

\bibitem{sener2018coreset}
Ozan Sener and Silvio Savarese.
\newblock Active learning for convolutional neural networks: A core-set
  approach.
\newblock In {\em International Conference on Learning Representations}, 2018.

\bibitem{activelearningsurvey}
Burr Settles.
\newblock Active learning.
\newblock {\em Synthesis Lectures on Artificial Intelligence and Machine
  Learning}, 6(1):1--114, 2012.

\bibitem{settles2014active}
Burr Settles.
\newblock Active learning literature survey. 2010.
\newblock {\em Computer Sciences Technical Report}, 1648, 2014.

\bibitem{shannon}
Claude~Elwood Shannon.
\newblock A mathematical theory of communication.
\newblock {\em Bell system technical journal}, 27(3):379--423, 1948.

\bibitem{vgg}
Karen Simonyan and Andrew Zisserman.
\newblock Very deep convolutional networks for large-scale image recognition.
\newblock {\em arXiv preprint arXiv:1409.1556}, 2014.

\bibitem{vaal}
Samarth Sinha, Sayna Ebrahimi, and Trevor Darrell.
\newblock Variational adversarial active learning.
\newblock In {\em Proceedings of the IEEE/CVF International Conference on
  Computer Vision (ICCV)}, October 2019.

\bibitem{prototypical}
Jake Snell, Kevin Swersky, and Richard Zemel.
\newblock Prototypical networks for few-shot learning.
\newblock In {\em Advances in neural information processing systems}, pages
  4077--4087, 2017.

\bibitem{springenberg2015unsupervised}
Jost~Tobias Springenberg.
\newblock Unsupervised and semi-supervised learning with categorical generative
  adversarial networks.
\newblock {\em arXiv preprint arXiv:1511.06390}, 2015.

\bibitem{mmen}
Chaofan Tao, Fengmao Lv, Lixin Duan, and Min Wu.
\newblock Minimax entropy network: Learning category-invariant features for
  domain adaptation.
\newblock {\em arXiv preprint arXiv:1904.09601}, 2019.

\bibitem{tong2001support}
Simon Tong and Daphne Koller.
\newblock Support vector machine active learning with applications to text
  classification.
\newblock {\em Journal of machine learning research}, 2(Nov):45--66, 2001.

\bibitem{pmlr-v97-tran19a}
Toan Tran, Thanh-Toan Do, Ian Reid, and Gustavo Carneiro.
\newblock Bayesian generative active deep learning.
\newblock volume~97 of {\em Proceedings of the 36th International Conference on
  Machine Learning (ICML)}, pages 6295--6304, Long Beach, California, USA,
  09--15 Jun 2019. PMLR.

\bibitem{tzeng2015simultaneous}
Eric Tzeng, Judy Hoffman, Trevor Darrell, and Kate Saenko.
\newblock Simultaneous deep transfer across domains and tasks.
\newblock In {\em Proceedings of the IEEE International Conference on Computer
  Vision}, pages 4068--4076, 2015.

\bibitem{adda}
Eric Tzeng, Judy Hoffman, Kate Saenko, and Trevor Darrell.
\newblock Adversarial discriminative domain adaptation.
\newblock In {\em Proceedings of the IEEE Conference on Computer Vision and
  Pattern Recognition}, pages 7167--7176, 2017.

\bibitem{inat}
Grant Van~Horn, Oisin Mac~Aodha, Yang Song, Yin Cui, Chen Sun, Alex Shepard,
  Hartwig Adam, Pietro Perona, and Serge Belongie.
\newblock The inaturalist species classification and detection dataset.
\newblock In {\em Proceedings of the IEEE Conference on Computer Vision and
  Pattern Recognition (CVPR)}, June 2018.

\bibitem{matchingnets}
Oriol Vinyals, Charles Blundell, Timothy Lillicrap, Daan Wierstra, et~al.
\newblock Matching networks for one shot learning.
\newblock In {\em Advances in neural information processing systems}, pages
  3630--3638, 2016.

\bibitem{wang2014label}
D. {Wang} and Y. {Shang}.
\newblock A new active labeling method for deep learning.
\newblock In {\em 2014 International Joint Conference on Neural Networks
  (IJCNN)}, pages 112--119, 2014.

\bibitem{minent}
Dequan Wang, Evan Shelhamer, Shaoteng Liu, Bruno Olshausen, and Trevor Darrell.
\newblock Fully test-time adaptation by entropy minimization.
\newblock {\em arXiv preprint arXiv:2006.10726}, 2020.

\bibitem{wang2017cost}
Keze Wang, Dongyu Zhang, Ya Li, Ruimao Zhang, and Liang Lin.
\newblock Cost-effective active learning for deep image classification.
\newblock {\em IEEE Transactions on Circuits and Systems for Video Technology},
  27(12):2591--2600, 2017.

\bibitem{active_erm}
Zheng Wang and Jieping Ye.
\newblock Querying discriminative and representative samples for batch mode
  active learning.
\newblock {\em ACM Transactions on Knowledge Discovery from Data (TKDD)},
  9(3):17, 2015.

\bibitem{suggestiveannotation}
Lin Yang, Yizhe Zhang, Jianxu Chen, Siyuan Zhang, and Danny~Z Chen.
\newblock Suggestive annotation: A deep active learning framework for
  biomedical image segmentation.
\newblock In {\em International Conference on Medical Image Computing and
  Computer-Assisted Intervention}, pages 399--407. Springer, 2017.

\bibitem{drn}
Fisher Yu, Vladlen Koltun, and Thomas~A Funkhouser.
\newblock Dilated residual networks.
\newblock In {\em CVPR}, volume~2, page~3, 2017.

\bibitem{bdd100k}
Fisher Yu, Wenqi Xian, Yingying Chen, Fangchen Liu, Mike Liao, Vashisht
  Madhavan, and Trevor Darrell.
\newblock Bdd100k: A diverse driving video database with scalable annotation
  tooling.
\newblock {\em arXiv preprint arXiv:1805.04687}, 2018.

\bibitem{zhu2017generative}
Jia-Jie Zhu and Jos{\'e} Bento.
\newblock Generative adversarial active learning.
\newblock {\em arXiv preprint arXiv:1702.07956}, 2017.

\end{thebibliography}
}

\clearpage
\appendix
\onecolumn
\section*{\centering {\LARGE{Supplementary Material}} \hfil}
\section*{\centering {\Large{Minimax Active Learning}} \hfil} 
\par
\section{Datasets Statistics}

Table \ref{tab:dataset_stats} shows a summary of the datasets used in our experiments. \cifar, \imagenet, and \inat are datasets used for image classification, while \city and \bdd were used for semantic segmentation. Budget for each dataset is the number of images that can be sampled at the end of each split. The size of the initial labeled pool and budget are given as percentage of the training set. 
\vskip -0.1in
\begin{table}[ht]\footnotesize
    \begin{center}
        \caption{Summary of the utilized datasets}
    \label{tab:dataset_stats}
    \vskip 0.1in
    \begin{tabular}{lcccccc}
        \hline
        Dataset & \#Classes & Train & Test & Initially labeled (\%) & Budget (\%) & Image size \\
        \toprule
        \cifar~\cite{cifar}& 100 & 50,000 & 10,000 & 2  & 2 & 32$\times$32 \\
        \inat~\cite{inat}& 8142 & 437,513 & 24,426  & 10 &  10 & 224$\times$224\\
        \imagenet~\cite{imagenet} & 1000 & 1,281,167 & 50,000  & 5 &  5 & 224$\times$224 \\
        \midrule
        \city~\cite{cityscapes} & 19 & 3475 & 2678 & 10 & 5  & 688$\times$688\\
        \bdd~\cite{bdd100k} & 19 & 7000 & 1000 & 10 &  5 & 688$\times$688\\    
        \bottomrule
    \end{tabular}
    \vskip 0.09in    
\end{center}
\end{table}

\end{document}